\documentclass[conference]{IEEEtran}
\IEEEoverridecommandlockouts
\usepackage{cite}
\usepackage{amsmath,amssymb,amsfonts}
\usepackage{algorithmic}
\usepackage{graphicx}
\usepackage{textcomp}
\usepackage{xcolor}
\usepackage[nolist]{acronym}
\usepackage{svg}
\usepackage{url}
\usepackage{diagbox}
\usepackage{balance}
\usepackage{multirow}
\usepackage[caption=false]{subfig}
\def\BibTeX{{\rm B\kern-.05em{\sc i\kern-.025em b}\kern-.08em
    T\kern-.1667em\lower.7ex\hbox{E}\kern-.125emX}}
\begin{document}

\title{Careful What You Wish For: on the Extraction of Adversarially Trained Models\\
\thanks{
% Identify applicable funding agency here. If none, delete this.
This research was funded by Synopsys Inc. and the Natural Sciences and Engineering Research Council of Canada (NSERC).
% \textcolor{red}{This research was funded by: TODO delete this part before blind review submission.}
}
}

\author{
% \IEEEauthorblockN{Anonymous authors}
% \IEEEauthorblockA{\textcolor{red}{TODO delete this part }\\
% \textcolor{red}{for blind review submission.} \\
% \textcolor{red}{for space estimation only}\\
% kacem.khaled@polymtl.ca}
% \and
\IEEEauthorblockN{Kacem Khaled}
\IEEEauthorblockA{\textit{Département GIGL} \\
\textit{Polytechnique Montréal}\\
Montreal, Canada \\
kacem.khaled@polymtl.ca}
\and
\IEEEauthorblockN{Gabriela Nicolescu}
\IEEEauthorblockA{\textit{Département GIGL} \\
\textit{Polytechnique Montréal}\\
Montreal, Canada \\
gabriela.nicolescu@polymtl.ca}
\and
\IEEEauthorblockN{Felipe Gohring de Magalhães}
\IEEEauthorblockA{\textit{Département GIGL} \\
\textit{Polytechnique Montréal}\\
Montreal, Canada \\
felipe.gohring-de-magalhaes@polymtl.ca}
% \and
% \and
% \and
}

\IEEEoverridecommandlockouts
\IEEEpubid{\makebox[\columnwidth]{To appear in Proceedings of PST 2022, Fredericton, Canada
\copyright2022
IEEE\hfill} \hspace{\columnsep}\makebox[\columnwidth]{}}

\maketitle

\newcommand\todo[1]{\textcolor{red}{#1}}

\begin{abstract}
Recent attacks on Machine Learning (ML) models such as evasion attacks with adversarial examples
and models stealing through extraction attacks 
pose several security and privacy threats.
Prior work proposes to use adversarial training to secure models from adversarial examples that can 
evade the classification of a model and deteriorate its performance.
However, this protection technique affects the model's decision boundary and its prediction probabilities, hence it might raise model privacy risks. 
In fact, a malicious user using only a query access to the prediction output of a model can extract it and obtain a high-accuracy and high-fidelity surrogate model.
To have a greater extraction, these attacks leverage the prediction probabilities of the victim model.
Indeed, all previous work on extraction attacks do not take into consideration the changes in the training process for security purposes. 
In this paper, we propose a framework to assess extraction attacks on adversarially trained models with vision datasets. To the best of our knowledge, our work is the first to perform such evaluation.
Through an extensive empirical study, we demonstrate that adversarially trained models are more vulnerable to extraction attacks than models obtained under natural training circumstances. 
They can achieve up to $\times1.2$ higher accuracy and agreement with a fraction lower than $\times0.75$ of the queries. 
We additionally find that the adversarial robustness capability is transferable through extraction attacks, i.e.,  extracted Deep Neural Networks (DNNs)
from robust models show an enhanced accuracy to adversarial examples compared to extracted DNNs from naturally trained (i.e. standard) models.

\end{abstract}

% Liste des sigles et abbréviations / List of acronyms and abbreviations
% \ifthenelse{\equal{\Langue}{english}}{
% 	\newcommand\abbrevname{LIST OF ABBREVIATIONS}
% }{
% 	\newcommand\abbrevname{LISTE DES SIGLES ET ABRÉVIATIONS}
% }
% \chapter*{\abbrevname}
% \addcontentsline{toc}{compteur}{\abbrevname}
% \pagestyle{pagenumber}
%
\begin{acronym}
  \acro{ML}{Machine Learning}
  \acro{DL}{Deep Learning}
  \acro{MLaaS}{Machine Learning as a Service}
  \acro{ANN}{Artificial Neural Networks}
  \acro{API}{Application Programming Interface}
  \acro{CNN}{Convolutional Neural Network}
  \acro{CNNs}{Convolutional Neural Networks}
  \acro{DNN}{Deep Neural Network}
  \acro{DNNs}{Deep Neural Networks}
  \acro{NN}{Neural Network}
  \acro{ART}{Adversarial Robustness Toolbox }
  \acro{DoS}{Denial-of-Service}
  \acro{FGSM}{Fast Gradient Sign Method}
  \acro{PGD}{Projected Gradient Descent}
  \acro{MLP}{Multilayer Perceptron}
  \acro{IP}{Intellectual Property}

\end{acronym}

\section{Introduction} % ~ 1 page
\label{sec:Introduction}
% Dans l'introduction, on présente le problème étudié et les buts
% poursuivis. L'introduction permet de faire connaître le cadre de la
% recherche et d'en préciser le domaine d'application. Elle fournit
% les précisions nécessaires en ce qui concerne le contexte de
% réalisation de la recherche, l'approche envisagée, l'évolution de
% la réalisation. En fait, l'introduction présente au lecteur ce
% qu'il doit savoir pour comprendre la recherche et en connaître la
% portée.
% \section{Introduction}\label{sec:Introduction}  

% \subsection{Context and Motivation}  
% While the first experiments with \ac{ANN} were conducted in the 1950s, \acf{DL} has only recently became widely used in crucial technology developments. 
% This breakthrough of \ac{DL} applications is due to the increase of dataset sizes as well as the availability of faster computational resources which helps running much larger models
% \cite{Goodfellow-et-al-2016}.
Given the success of \ac{DL} in achieving the state-of-the-art and sometimes human-competitve performance in several computer vision tasks
\cite{krizhevsky2017imagenet, le2011learning, taigman2014deepface}, %TODO cite resources
\ac{DL} became the core of several critical applications such as autonomous vehicles and robotics. 
\acf{ML} providers offer their fine-tuned models to users as a service (\ac{MLaaS}) so they can benefit from its outstanding performance in a specific prediction task. 
The architecture of the models is usually not revealed, instead, users are only provided with an \ac{API} to query them \cite{papernot2017practical}. % CM-any citations? 

Models are often proprietary and are a business advantage to their owner as they are expensive to obtain~\cite{strubell2019energy}. 
In fact, when a model owner shares an internal knowledge about a model such as the architecture or weights, a malicious user can infer sensitive information in the training data which raises privacy concerns in certain cases~\cite{shokri2017membership}. 
Additionally, revealing the trained models poses security concerns, since an adversary can craft adversarial examples evading the classification by a white-box victim model easily 
\cite{szegedy2013intriguing}.

Prior work shows that an adversary with a query access to a model is able to extract it and construct a very similar model to the victim's functionality 
\cite{tramer2016stealing, papernot2017practical,correia2018copycat ,orekondy2019knockoff, Krishna2019, Reith2019, Truong2020, Pal2020, Yuan2020, Jagielski2020, yuan2020attack, kariyappa2020maze, chawla2021data}. %TODO cite attacks
These attacks help an adversary gain more knowledge about the victim model and therefore raise different concerns;
for example, using an extraction attack, an adversary can craft adversarial examples evading the extracted model that are transferable for the evasion of the original one 
\cite{papernot2017practical}.
% In the case of a \ac{NN} accelerator, an access to side-channels such as power or memory is used by certain adversaries to extract the model
% \cite{Wei2018sca, Yan2018sca, Hong2018sca, Yoshida2019sca,Batina2019sca,Duddu2019sca, Hu2020sca, Wei2020sca, Hong2020sca, Liu2020sca, Xiang2020sca}. %TODO cite attacks
%CM- why in contrast ?? there is no contradiction
In contrast, recent work shows attempts to defend against these extraction attacks and mitigate their risk 
\cite{Kesarwani2018,lee2018defending,Juuti2019,kariyappa2020defending,Orekondy2019poisoning}. %TODO cite defenses
They often work on slightly changing the output prediction values without highly affecting
the model's performance. This may hinder the possibility of an attack, or at some points make it require more queries to achieve a sufficient extraction. %CM- citation ? 

In addition to the extraction attacks that compromise the privacy of models, 
the \ac{ML} security gained significant attention during the last decade especially with the serious threats of adversarial attacks \cite{papernot2016towards}. 
Recently, model owners find themselves required to modify their  regular training process that results in unprotected models referred also as \textit{natural} models. They apply a defense technique to their \ac{DNNs} such as \textit{adversarial training} before deploying them as final oracles~\cite{madry2017towards}. 
%CM-work on below-clarify the gap/problem
Nevertheless, existing extraction attacks do not take into consideration these security-imposed and uncommon scenarios that deviate from the normal training process, and this is still, to the best of our knowledge, an open research problem. 
Since researchers tend to work on different attack objectives separately and evaluate different potential threats independently, this may raise concerns about the robustness under these circumstances. 
Hence, we don't know how the risk of model extraction would develop in the case of adversarial training and whether models get more or less vulnerable to these attacks. 
As a matter of fact, a recent work showed that when working on improving the security of a model, they may have increased the risk of the training data's privacy and leaked information about it~\cite{song2019privacy}.

In this paper, 
% we bridge the gap between the current state-of-the-art research in \ac{DL} security and privacy and an open research question. 
we explore important intersections between extraction attacks and an unavoidable security-imposed scenario which is protecting natural models (i.e., unmodified models) against adversarial examples.
% ------------------------------------
% subsection Contributions
% ------------------------------------
% In this paper, 
We assess the robustness against extraction attacks of \ac{DL} models under different learning circumstances.
Thus, we make the following contributions:
\begin{itemize}
\item an open source framework to assess a model’s extractability against extraction attacks under different unexplored situations in the state-of-the-art using potential defenses against adversarial attacks\footnote{Our code is available at the GitHub repository: 
% \url{
https://github.com/KacemKhaled/model-stealing
% }
% Link omitted for blind review.
}; % enabling the completion of a full risk assessment.
%Quantitative measurement of attacks: Is it possible to measure the effect of attacks? If so, we could compare the security performance of learning algorithms. We could calculate risk based on probability and damage assessments of attacks. enabling a full risk assessment to be completed.
\item an evaluation of the extraction risk of models with increased robustness against adversarial examples showing that adversarially trained models can achieve up to $\times1.2$ higher accuracy and agreement (i.e., fidelity) of natural models with a fraction lower than $\times0.75$ of the queries, and;
\item an empirical proof that the adversarial robustness capability of adversarially trained models is transferable through models extraction attacks.  

\end{itemize}

%impact
% this may be a conclusion  %CM i feel too general-focus security/privacy -- immediate impact of your work 

% An extraction attack poses several threats depending on the adversary's purpose. For example, thieves might endanger the model owner’s business through providing the stolen model to competitors or that might use the \ac{ML} service without any cost. 
% In addition, stealing the functionality of a model can boost other attacks that rely on a wider knowledge of the model (e.g. evasion attacks), therefore, protecting the \ac{DNNs} against extraction attacks may close the door in front of other potential attacks or decrease their risks.

% \textcolor{red}{Structure of the Paper}
%  TODO add labels to sections instead of numbers
The remainder of the paper is organized as follows: 
Section~\ref{sec:ML-DL} introduces the basic notions to \ac{ML} and \ac{DL}; 
Section~\ref{sec:background-sota} reviews the background and the state-of-the-art that relates to privacy and security threats and attacks; 
Section~\ref{sec:methodology} details our methodology; 
we include our experiments and obtained results in Section~\ref{sec:experiments}; 
Section~\ref{sec:defenses} summarizes potential defenses and countermeasures against extraction attacks;
% we review defenses to extraction attacks in Section~\ref{sec:defenses};
and Section~\ref{sec:conclusion} concludes this paper.
%, including our future plans.

% % \section{Motivation}  % ~  < 0.5 page

\section{Machine Learning and neural networks}  % ~ 2 pages
\label{sec:ML-DL}
% part for literature review
% + part to introduce ML and DL

% \subsection{Machine learning and neural networks}

%\textcolor{red}{TODO : remove the ML section and talk faster about DL}

\textit{A machine learning} algorithm
is an algorithm that can learn from data to perform a task. The data includes examples with quantitatively measured features \cite{Goodfellow-et-al-2016}. 
An example is typically represented as a vector 
$ \boldsymbol x \in \mathcal{R}^n$ 
where each entry of the vector $x_i$ is another feature.
\ac{ML} algorithms can be broadly categorized as supervised and unsupervised:

% TODO : Supervised and unsupervised learning ??
\begin{itemize}

\item supervised learning algorithms involve working on annotated datasets which means every example $\boldsymbol x$ in the dataset is associated with a  provided label $\boldsymbol y$. The algorithm learns to predict $\boldsymbol y$ from $\boldsymbol x$ by estimating $p(\boldsymbol y|\boldsymbol x)$~\cite{Goodfellow-et-al-2016}, and;
\item unsupervised learning algorithms experience an unlabeled dataset containing a collection of examples, and try to learn interesting properties about the dataset's structure~\cite{Goodfellow-et-al-2016}. The algorithm wants to learn implicitly or explicitly a probability distribution that generated the data $p(\boldsymbol x)$.

\end{itemize}
Other \ac{ML} algorithms do not work on just fixed data, such as reinforcement learning algorithms where the learning system interacts with a dynamic environment in which it must perform a certain goal. The system is provided with a feedback loop in terms of rewards as it takes actions in the environment, and through experience the agent learns a policy that maximizes that reward\cite{Russell2003}.
% \todo{ Semi-supervised and self-supervised learning, GANs...?}

\begin{figure}[tbp]
\centering
\includegraphics[width=0.49\textwidth]{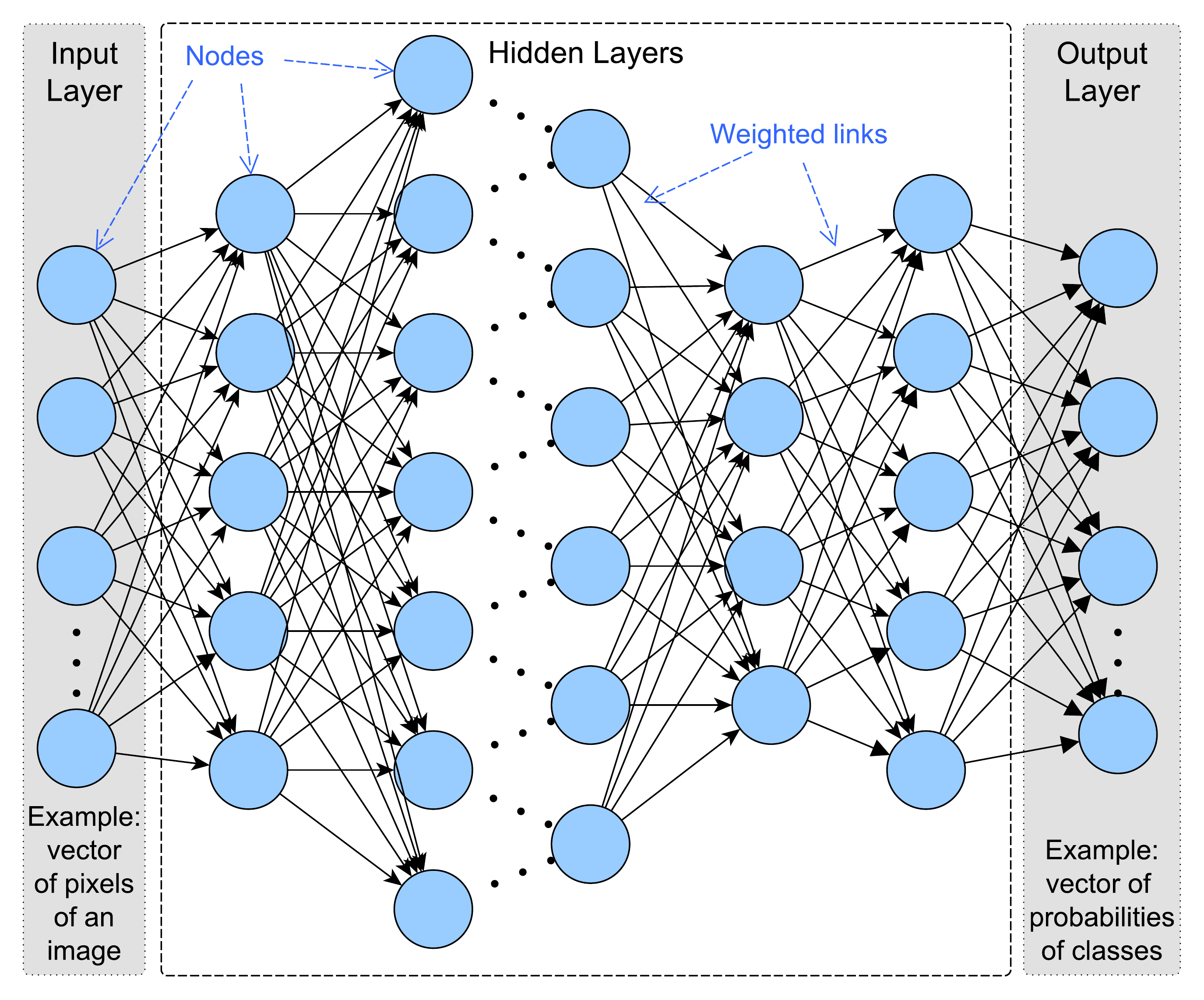}
\caption{An illustration of a neural network with multiple layers. The input values are fed to the first layer and propagated to the following layers multiplied by weights. For each layer, the nodes output the results of an activation function to the sum of their entries. The last layer outputs probabilities for each class at the corresponding node.}
\label{fig:neural net}
\end{figure}

\textit{Deep learning} is a subset of \ac{ML} involving techniques based on \ac{ANN} 
\cite{Goodfellow-et-al-2016}. 
An example of a deep learning model is a feed-forward deep network, or a \ac{MLP}, which is a mathematical function defined as 
$\boldsymbol y = F(\boldsymbol x)$ 
mapping input vectors 
$\boldsymbol x \in \mathcal{R}^n$ 
to output  values 
$\boldsymbol y \in \mathcal{R}^m$. 
A \ac{NN} has $L$ layers characterizing the model's depth where each layer has a width and is characterized with a number of nodes called neurons. 
Fig~\ref{fig:neural net} illustrates an example architecture where the neurons connect the layers together through weights $\boldsymbol W$. Each neuron perform an activation function $g(\boldsymbol z)$ to the weighted sum of the output values of the previous layer's neurons: $\boldsymbol z= \boldsymbol W^T\boldsymbol h+\boldsymbol b$.

To obtain a meaningful set of weights, a training process is performed and the parameters are optimized via approaches such as stochastic gradient descent (SGD) in order to minimize a certain loss function~\cite{Goodfellow-et-al-2016}.

\textit{\ac{CNNs}}
are a particular type of feed-forward \ac{NN}s, that are mostly suitable for image classification tasks, with the requirement that at least one of the layers performs a mathematical operation called convolution operation followed by a non-linear activation. 
After enough training, \ac{CNNs} are able to learn filters and characteristics relevant to the task \cite{Krizhevsky2017}.

\section{Background and related work} 
\label{sec:background-sota}
% \subsection{Threat model of attacks on machine learning}
Recently, \ac{ML} models including \ac{DNNs} were shown to be vulnerable to several attacks \cite{Barreno2006, Biggio2012, Biggio2013, szegedy2013intriguing, Papernot2016limitations}. In this section we taxonomize these attacks depending on the adversary's goals and capabilities as well as the underlying assumptions of his knowledge about the victim.

%\subsubsection{Threat Model}

\subsection{Adversary's capability}
It defines the actions that the attacker could perform against the victim model. These capabilities could be related to the training phase or the testing phase depending on the attacker strategy and purposes.
In the training phase, the attacker may be able to inject new data to the training set, or add malicious modifications to the current one.
In the testing phase, the attacker can only control the test data, for example inject well crafted examples to the model in order to lead it to misclassification.

\subsection{Adversary's knowledge}
Models could be affected by different attacks depending on the knowledge level of the adversary which reflects the strength of the attack. Taking the attack settings into consideration, we can cite three types of attacks: 
\begin{itemize}
\item {\textit{White-box attacks:}} 
the adversary has full access and knowledge about the model (e.g. network architecture, parameters, training set...);

\item {\textit{Gray-box attacks:}} 
the adversary assumes a partial knowledge about the targeted model (e.g. the distribution of the training set, the classifier type), and;

\item {\textit{Black-box attacks:}}
the attacker has a very limited access to the target model. He can observe the inputs and outputs, but has no access to the internal working of the model neither the training set.
\end{itemize}

% ---

\subsection{Adversary's goals (attacks on machine learning models)}
The attacker can exploit several parts of the machine learning pipeline to achieve various malicious goals \cite{Goodfellow2018darkarts}. Attacks can be performed in the training phase or during the inference phase. 
We present the different threats depending on the adversary's objective.
% \begin{figure}[tb]
% \centering
% \includegraphics[width=4in]{Attacks pipeline}
% \caption{Attacks on the machine learning pipeline   \cite{Goodfellow2018darkarts}}
% \label{fig:Attacks pipeline}
% \end{figure}

\subsubsection{Poisoning attacks}
refer to causative attacks in which specially crafted attack points are injected into the training data in order to increase the model’s test error \cite{Biggio2013} and corrupt it. \emph{“Causative attacks alter the training process through influence over the training data”} \cite{Barreno2006}. Poisoning attacks assume that the attacker knows the learning algorithm and can draw points from the data distribution.
Finding such an attack point can be formulated as an optimization problem. In fact, for a poisoning attack against \ac{ML} model, the adversary’s goal is to find an attack point whose addition to the training set maximally decreases the model’s classification error \cite{Biggio2012}. 
Poisining attacks were demonstrated against several applications such as spam filters, PDF malware detection, \ac{DoS} attack detection, handwritten digits recognition and healthcare~\cite{nelson2008exploiting, xiao2015feature,rubinstein2009antidote,jagielski2018manipulating}. 

\subsubsection{Membership inference attacks}
in these attacks, the adversary aims to determine whether an individual data record is part of the training set of the attacked model. 
This attack reflects the leaked information about the training set from the model \cite{song2019privacy}.
This attack leverages the inference part of the model, so it does not interfere with the training phase, but instead the adversary uses test data in order to achieve his goal. 
This attack poses a high risk on the privacy of the data when sensitive information is used to train the model (e.g. medical records). In some cases, this attack relies on a confidence thresholding technique, in which the adversary concludes about a membership of data through the prediction confidence given by the model about this input.

% Defences that mitigate these attacks include temperature scaling which reduces the difference between the prediction confidence of the training model on its training set and test set. Another way is to use the regularization in order to improve the robustness generalization of the model.

\subsubsection{Evasion attacks (i.e., adversarial examples attacks)}

this attack happens on the inference part of the model, so it does not interfere with its training phase. Instead, it interferes with the test data through creating adversarial examples.  
\emph{“An adversarial example is a sample of input data which has been modified very slightly in a way that is intended to cause a machine learning classifier to misclassify it”}\cite{Kurakin2017}. 
Usually these modifications are not subtle to humans, but they cause a huge error in machine learning models and it was shown that even state-of-the-art algorithms can be fooled by this malicious input. 
The methods of crafting adversarial examples rely on optimization problem that aims to maximise the networks prediction error \cite{szegedy2013intriguing}. 

Papernot et al. \cite{papernot2016crafting} demonstrated that these adversarial attacks against feed-forward neural networks can be adapted to fool Recurrent Neural Networks (RNNs). They particularly demonstrated an attack against the Long-Short-Term-Memory (LSTM) architecture. 
Adversarial attacks were also shown to be applied in the physical world, Kurakin et al. \cite{kurakin2016adversarial} proved that adversarial images printed on paper can be misclassified when they are fed to a classifier through a cellphone camera. 
Furthermore, Eykholt et al. \cite{eykholt2018robust} explored the use case of safety-critical situations where road signs could be manipulated to fool the autonomous vehicles classifiers. They succeded in proposing physical perturbations such as stickers and graffiti to road signs that were misclassified by the target classifier in lab settings as well as in field test with captured video frames obtained from a moving vehicle.

Several countermeasures has been proposed to mitigate these attacks, for example adversarial training, which rely on creating a bigger dataset rich with adversarial examples through generating adversarial examples using the methods that could be used by the attacker and then, retrain the model using that dataset \cite{madry2017towards , song2019privacy}. Another way to defend against adversarial examples is through modifying the model with network distillation
\cite{Papernot2016distillation} which relies on transferring the knowledge from an initial network to a distilled network.

\subsubsection{Model extraction attacks (model stealing attacks)}
in these attacks the adversary exploits the inference phase of the machine learning pipeline. Through observing the prediction outputs of his queries, even in black-box settings, the attacker aims to steal the functionality  or parameters of a \ac{ML} model~\cite{tramer2016stealing}.
The adversary attempts to learn a classifier $\hat{f}$ that matches or closely approximates a target classifier $f$. These attacks have a great risk on the \ac{IP} of model owners. Additionally, they facilitate other attacks such as membership inference and evasion attacks.

In the literature, extraction attacks can broadly be categorized as: 
\begin{itemize}
    \item attacks that exploit hardware access, or side-channel extraction attacks: when a \ac{ML} model is deployed in a hardware platform where the user has access to it, for example on an FPGA, on a \ac{NN} accelerator or sometimes the same host machine as the victim, the adversary has more access than the API software level case. Leveraging the leaked information through side-channels (e.g. cache, memory, power, electromagnetic, timing), the adversary can gain more knowledge about a \ac{NN} architecture, its parameters, or even duplicate its functionality \cite{Wei2018sca, Yan2018sca, Hong2018sca, Batina2019sca, Hu2020sca, Hong2020sca, Liu2020sca, Xiang2020sca}, and; %TODO cite attacks

    \item attacks that leverage \ac{API} query access, through exploiting the input-output predictions on a software level~\cite{tramer2016stealing, papernot2017practical,correia2018copycat ,orekondy2019knockoff, Krishna2019, Reith2019, Truong2020, Pal2020, Yuan2020, Jagielski2020}. These attacks are the main focus of our paper, therefore, we describe them in more details in the next section.
    % along with the state-of-the-art defenses that mitigates their risk.
\end{itemize}

\begin{figure*}[htbp]
\centering
\includegraphics[width=0.85\textwidth]{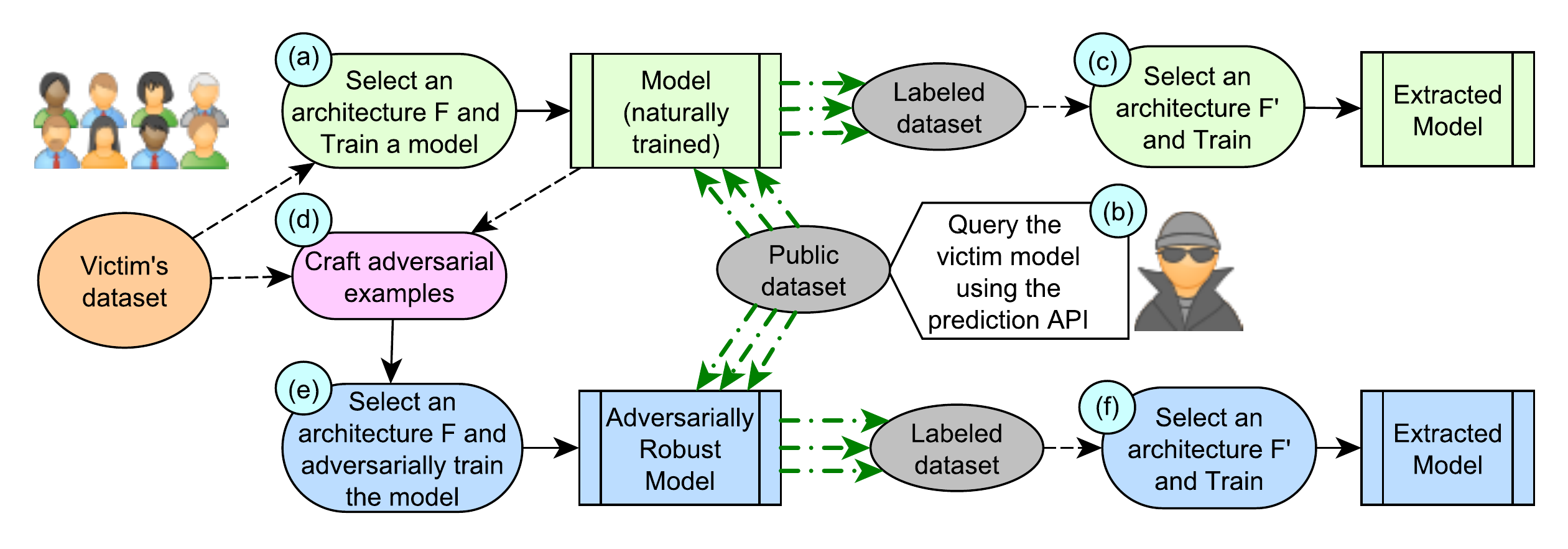}
\caption{Evaluation of the extraction risk of natural models versus adversarially trained models.
First, model owners train an oracle under natural training circumstances (a) and offer their model for queries with an \ac{API}. 
The adversary labels a public dataset through observing the queries responses from the victim model (b) which he uses later to train a surrogate, called extracted model (c). 
Alternatively, a model owner generates adversarial examples (d) to adversarially train their model (e) which the adversary would extract using the same process (b), (f).
}
\label{fig:experiment}
\end{figure*}

\subsubsection{API based extraction attacks}

% \begin{figure}[htb]
% \centering
% \includegraphics[width=4in]{dia/tramer model stealing.pdf}
% \caption{Diagram of ML model extraction attacks using API queries \cite{tramer2016stealing} }
% \label{fig:Tramer model stealing}
% \end{figure}

% \textcolor{red}{TODO: maybe summarize this part to keep 50\%}
Jagielski et al. \cite{Jagielski2020} taxonomize model extraction around the two adversarial objectives: \emph{accuracy} and \emph{fidelity}. Accuracy measures how well the extracted model is performing on the underlying learning task and the goal is to extract a model that tries to make accurate predictions. Fidelity measures the matching between the predictions of the extracted model and the victim model on any input.

% A Survey of Privacy Attacks in Machine Learning, MARIA RIGAKI

Initial works on stealing attacks have been demonstrated by Lowd \& Meek \cite{lowd2005adversarial} where they extract linear classifiers suchs as support vector machines (SVM) with linear kernels and logistic regressions (LR). 
They assume that the adversary has a black-box access to the oracle and the queries return just the predicted class label.
Tramer et al. \cite{tramer2016stealing} propose more attack techniques that works on simple linear and non-linear architectures such as SVMs, LRs, decision trees and simple neural networks. 
They present the scenario of \ac{ML} model stealing where a data owner has a trained model $f$ and allows others to make prediction queries, an adversary uses $q$ prediction queries to extract an $\hat{f} \approx f$.
Some of their attacks are based on equation-solving to find the parameters of a target model relying on the queries. 
They also suggest a path-finding algorithm to extract decision trees. 
%As an adversary dataset, they use random uniform noise samples 

Papernot et al. \cite{papernot2017practical} propose an extraction attack technique that facilitates crafting transferable adversarial examples fooling the target victim. 
The idea of their attack is to select a substitute \ac{DNN} architecture to the attacked \ac{DNN} and train it in a way that imitates the target model using synthetic data generation and labeling the data with the target model. 
They assume that the attacker has a black-box capability, but initially draw a dataset from the same distribution to query the target model, then using a Jacobian-based data augmentation approach, they find examples defining the decision boundary of the target model. 
Finally, using these labeled examples they obtain a high-fidelity surrogate model. Then, they craft adversarial examples that are transferable for evading the classification on the original target model.
Juuti et al. \cite{Juuti2019} present a concurrent framework which relies on synthetic data generation with Jacobian-based data augmentation and randomly perturbing color channels. They further investigate selecting hyperparemeters for the surrogate model instead of using fixed ones.

% Oh et al. \cite{oh2019towards} extend model stealing to infer internal information about the target model through using a "metamodel" that is trained on a diverse set of white-box models ("meta-training set") close to the attacked black-box victim. Their metamodel takes a model's query input-output pairs as input and returns the corresponding model attributes as output, i.e. which architecture, optimisation process and which training dataset.
% Wang \& Gong \cite{wang2018stealing} investigate stealing hyperparemeters when the attacker knows the training dataset and the \ac{ML} algorithm.  They demonstrate their attack on ridge regression, SVMs and \ac{NN}s.

Correia-Silva et al. \cite{correia2018copycat} demonstrate that a \ac{CNN} can be copied using public data to query the target model. Their technique \emph{"Copycat CNN"} leverages a mix of problem domain and non-problem domain data. They follow the same concept of generating a fake dataset labeled by the target network, then use it to train the copycat network. They successfully extracted models on problems such as facial expression, object and crosswalk classification.
Orekondy et al. \cite{orekondy2019knockoff} propose \emph{"Knockoff Nets"} attack where the adversary is lacking knowledge about the train/test data used by the target model and its internal architecture. 
They assume that the attacker is only capable to interact with the victim model through querying it with images and observing the predictions. Like previous work, their attack is based on querying the target model with samples and contructing a dataset which serves to train the knockoff model.
They investigate more complex \ac{DNNs} and rely on publicly available datasets to steal the victim model such as ILSVRC \cite{deng2009imagenet , ILSVRC15} and OpenImages \cite{kuznetsova2020open}.

In addition to computer vision tasks, model extraction is also effective on natural language processing (NLP) tasks. Krishna et al. \cite{krishna2019thieves} demonstrate an extraction attack against pretrained and fine-tuned large language models such as BERT \cite{devlin2018bert}. The attacker uses the same pretrained model as the one assumed to the victim and fine-tunes it on his obtained fake dataset. In order to build this dataset, the adversary queries the target model with random sequences of words coupled with task-specific heuristics and does not need to have grammatical nor semantical meaning, and then uses the victim outputs as labels.

Recently, other works in extraction attacks focus on the techniques used to generate the adversary's dataset instead of relying on public datasets \cite{han2020robustness, chawla2021data}. In addition to synthetic data generation and data augmentation, they leverage generative models to generate data with an objective that enables a better extraction~\cite{kariyappa2020maze}.

\subsection{Positioning with the state-of-the-art} 
% Explain here what is original in this paper
% The extraction attacks raise risks on the intellectual property of \ac{ML} models and it may facilitate other adversarial attacks. 
Stealing attacks are diverse and were proven to be successful in many cases, yet they were only demonstrated on unaltered and standard sate-of-the-art models that were trained under natural training circumstances. 
They still do not consider security-imposed scenarios that modify the model's architecture or its behavior for instance towards adversarial examples. 
Adversarial training impacts the prediction probabilities that could be leveraged by malicious users to perform other attacks through observing the model's response to their queries.
Particularly, adversarially robust models were proven to be more vulnerable to membership inference attacks due to their wider generalization compared to natural models which raises concerns about the training data privacy~\cite{song2019privacy}. 
Extraction attacks rely on the predictions probabilities of the victim model to the adversary's queries and have not yet been evaluated on adversarially robust models, hence, the extraction robustness of the latter is still an open research problem.
It is unknown if we alter the models through adversarial training, we increase or not their vulnerability to extraction.
% Our paper seek to lessen this research gap with a new approach and significant empirical study. 
In order to answer to this question, we propose a new methodology and a significant empirical study.

\section{Methodology} % ~ 2.5 pages
\label{sec:methodology}
% background of terms or concepts directly related to the paper
% + theory stuff
% ex: attack taxonomy, attack methodology
% \input{paper_body/Background of the paper}

% \subsection{Methodology}
 Consider a victim classifier trained solely on benign (i.e., natural) examples and another victim trained on both benign and adversarial examples to increase its adversarial robustness. We call the former \emph{Natural model} and the latter \emph{Adversarially (Adv.) trained model}. 
%  The research question we seek to answer is which one is more vulnerable to extraction attacks, the \emph{Natural model} or the \emph{Adv. trained model}?

\subsection{Threat model}
In our work, we consider the state-of-the-art extraction attack KnockoffNets \cite{orekondy2019knockoff} that aims to steal the functionality of black-box model trained with an architecture $F$ through leveraging the API access.
Fig.~\ref{fig:experiment}  gives an overview of our methodology.
Following the attack strategy, we query a victim, trained on a dataset $D_V$, with a different dataset $D_A$ and through observing the obtained prediction probabilities for each sample, we obtain a labeled \emph{transferset} (Fig.~\ref{fig:experiment} (a), (b)). Then, we select an architecture $F'$ to train the surrogate model leveraging the probabilities as soft labels for the adversary's dataset $D_A$ (Fig.~\ref{fig:experiment} (c)).
The attack is performed in black-box settings, i.e., the adversary has no prior knowledge about the victim's architecture nor the training set.

\subsection{Extraction of Adversarially trained models}

We adopt different state-of-the-art techniques to generate adversarial examples, namely \ac{FGSM} and \ac{PGD}: 

\begin{figure}[]
\centering
\includegraphics[width=0.48\textwidth]{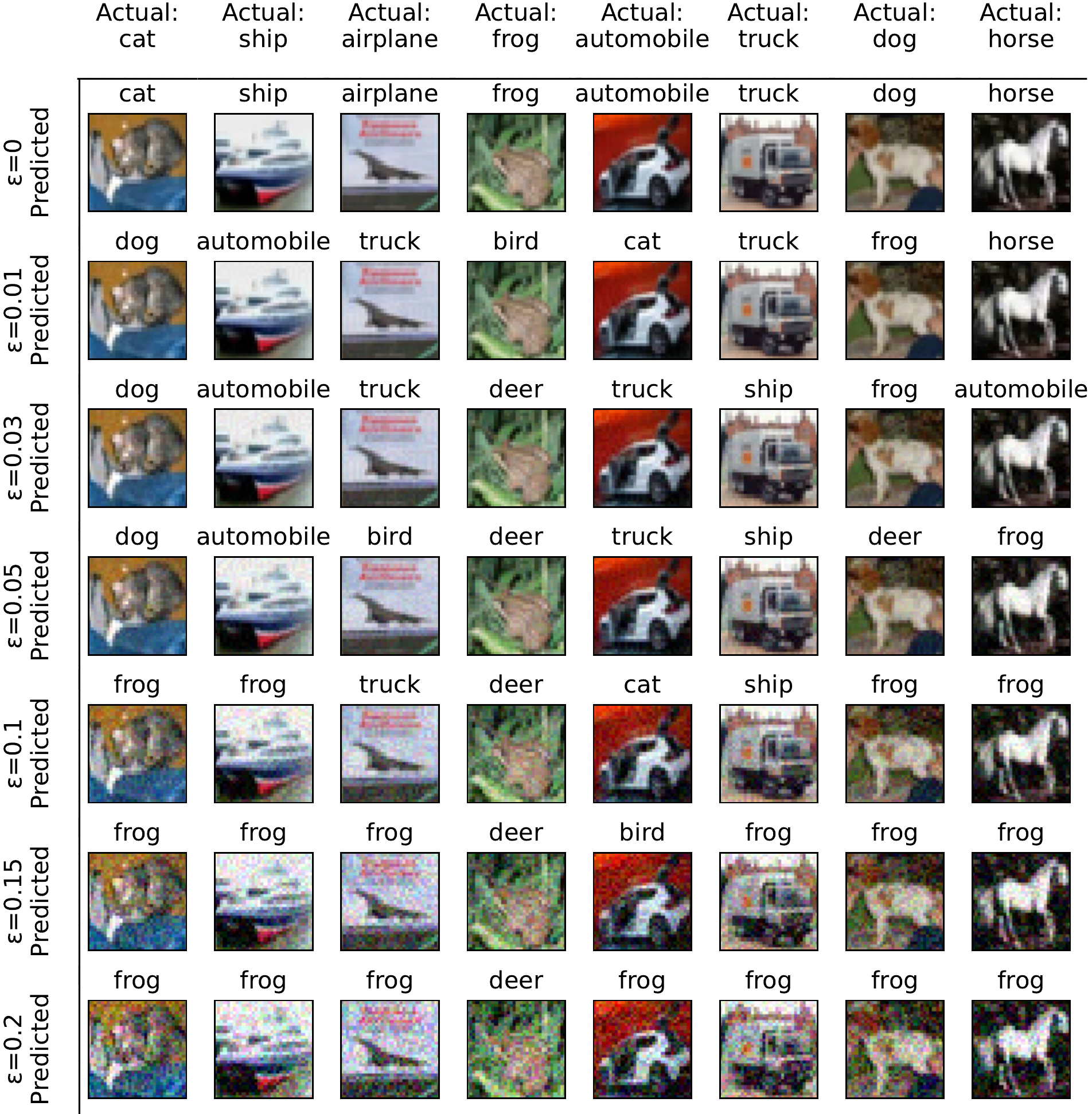}
\caption{A demonstration of PGD evasion attack on a ResNet-34 architecture trained on CIFAR-10 dataset. The header shows the actual ground truth labels. Then, for each row, we visualize a set of the resulting adversarial images and their prediction labels with relation to the attack perturbation level $\varepsilon$. For lower $\varepsilon$ values, the attack is imperceptible, yet effective in changing the predicted labels. The higher the perturbation level $\varepsilon$, the more visible the distortion on the adversarial examples.}
\label{fig:adversarial cifar10}
\end{figure}

% \subparagraph{\textit{Adversarial examples generation:}}
% In our paper, we use two techniques from the state-of-the-art to generate adversarial examples: 
\begin{itemize}
    \item \ac{FGSM} \cite{goodfellow2014explaining} uses the gradient of the loss with respect to the input image to craft new adversarial image that maximizes the loss. 
    \begin{equation}
    x_{adv} = x + \varepsilon sign( \nabla_x L(\theta,x,y) )
    \label{eq:FGSM}
    \end{equation}
    The equation \ref{eq:FGSM} summarizes this technique, where $x_{adv}$ is the adversarial image, $x$ the input image, $y$ the input label, $\varepsilon$ a parameter to control the perturbations amplitude, $\theta$ the model parameters, $L$ the loss of the trained model and $\nabla_x$ the gradient with respect to $x$.

    \item \ac{PGD} \cite{madry2017towards} is an iterative method that is considered as a multi-step variant of the previous technique, which is a projected gradient descent on the negative loss function. The PGD attack contains $T$ gradient descent steps. 
    \begin{equation}
    x^{t+1} = \Pi_{x+S} (x^{t} + \varepsilon sign( \nabla_x L(\theta,x,y) ) )
    \label{eq:PGD}
    \end{equation}
    Equation \ref{eq:PGD} summarizes this technique, where $\Pi_{x+S}$ denotes the projection onto the perturbation constraint $x+S$ and $S$ is the set of allowed perturbations. 

\end{itemize}
% TODO : here you can exlpain more evasion attacks if needed

We demonstrate in Fig.~\ref{fig:adversarial cifar10} a PGD attack that generates an adversarial example for each image by adding an imperceptibly small vector. The resulting malicious input changes the prediction of each image by ResNet-34~\cite{he2016deep}, one of the state-of-the-art classifiers for images.  

Leveraging the generated adversarial examples from either the PGD technique or the FGSM technique, we expand the training set and retrain the models (Fig.~\ref{fig:experiment} (d), (e)). 
We obtain adversarially robust models. We evaluate the robustness of obtained models against adversarial examples generated with both techniques.
Finally, we evaluate the extraction risk of these obtained adversarially robust models (Fig.~\ref{fig:experiment} (b), (f)).
\subsection{Evaluation metrics}
To evaluate the success of extraction attacks, we calculate the \emph{accuracy} of the surrogate model on the victim's test set, which is the fraction of correct predictions from all predictions. This metric measures the stolen functionality from the victim, i.e., how well the surrogate model performs on the same task. 
In addition, we measure the \emph{agreement} (i.e., fidelity) between the victim model and the extracted model through calculating the fraction of matching predictions (both correct and wrong) from the victim and the surrogate model, this is similar to calculating the accuracy of the extracted model using the predictions of the attacked victim as ground truth labels. 
Both metrics are calculated using a heldout labeled test set that was not seen before by the victim (during the training) nor the extracted model.

% The proposed metric is indicated in equation \ref{eq:acc_adv}. The higher this extraction rate, the more successful the attack.

% \begin{equation}
% Accuracy = \frac{Nb.\; of\; correct\; predictions}{Total\;nb.\;of\;predictions}
% \label{eq:acc_adv}
% \end{equation}

% \begin{equation}
% Agreement = \frac{Nb.\; of\; similar\; predictions\; as\; victim\; model}{Total\;nb.\;of\;predictions}
% \label{eq:agreement}
% \end{equation}

% \begin{equation}
% Extraction\,Rate = \frac{Accuracy\; of\; the\; extracted\; model}{Accuracy\;of\;the\;victim\;model}
% \label{eq:ER}
% \end{equation}
Then, we evaluate the success of extraction attacks against \emph{Adv. trained models}
versus \emph{Natural models} and study whether making the models robust against adversarial examples makes them more vulnerable to extraction attacks. 
Finally, we measure the robustness of victims and surrogates against evasion attacks. We compute the \emph{adv. accuracy} of each model against various adversarial examples generated with FGSM and PGD techniques using multiple levels of perturbations  $\varepsilon$.

% For instance, in case some adversarially robust models are less vulnerable to extraction attacks than natural (i.e. default) models, we could say that we discovered a hidden advantage of a twofold technique that defends against both adversarial examples and extraction attacks. Furthermore, adapting these techniques and exploring other ones may advance the state-of-the-art defenses against extraction attacks.
% In the other case, if we find out that some adversarially robust models are more vulnerable to extraction attacks, then our research will as well discover a serious problem that model owners should be aware of. These potential threats should be avoided and their risk should as well be mitigated.

\section{Experiments}
\label{sec:experiments}
\subsection{Setup}
% \Chapter{ILLUSTRATIVE EXAMPLE AND PRELIMINARY RESULTS}\label{sec:Example}

% In this chapter, we describe our progress in the research project and present the preliminary results that we found so far.

% \section{Experiments}

\begin{table}[tp] %[htbp]
  \centering
  \caption{Selected datasets in our experiments. The first three rows are the victim's datasets and the following rows are the adversary's datasets.}
%   \begin{tabular}{|p{1.5cm}|p{3.3cm}|p{1.8cm}|p{2.6cm}|p{2.3cm}|}
%   \begin{tabular}{|p{1.2cm}|p{2.7cm}|p{0.8cm}|p{1.2cm}|p{0.8cm}|}%p{2cm}|
  \begin{tabular}{|p{1.2cm}p{2.7cm}p{0.8cm}p{1.2cm}p{0.75cm}|}%p{2cm}|
    \hline
    % \rowcolor[gray]{0.8}\color{black}
    \textbf{Dataset}    &   \textbf{Description}  & \textbf{Image size} &  \textbf{Nb. of samples}  & \textbf{Nb. of classes} %& Samples 
    \\ \hline \hline
    MNIST   &  Handwritten grayscale digits & 28$\times$28 & Train: 50k \newline   Test: 10k & 10 %& TODO   
    \\ \hline
    CIFAR-10   &  Images of animals and vehicles & 32$\times$32 & Train: 60k  \newline Test: 10k  & 10 %& TODO  
    \\ \hline
    SVHN   &  Images of street view house numbers & 32$\times$32 & Train: 73k \newline Test: 26k  & 10 %& TODO  Train: 73257 Test: 26032
    \\\hline %& TODO
    \hline
    Fashion MNIST   &  Grayscale images of clothes & 28$\times$28 & Train: 50k \newline   Test: 10k & 10 %& TODO
    \\ \hline
    ImageNet  &  Various real-world images belonging to 1000 different categories & 224$\times$224 &  Train: 1.2M \newline Test: 50k  & 1000
    % \\\hline
    % GTSRB   &   A german traffic sign dataset & 32x32 &  39209 / 12630  & 43 %& TODO   
    % \\\hline
    % CUB-200   &  Images of 200 bird species & 224x224 & 5994 / 5794  & 200 %& TODO   
    % \\\hline
    % Caltech-256  &  Various real-world images belonging to 256 different categories & 224x224 &  23703 / 6904  & 256 %& TODO   
    \\\hline
    
  \end{tabular}
  \label{tab:Datasets}
\end{table}

In our experiments we tackle \ac{DNNs} trained on 3 benchmark images datasets: MNIST 
\cite{lecun2010mnist}, CIFAR-10 \cite{krizhevsky2009cifar10} and SVHN~\cite{netzer2011svhn}. We describe these datasets as well as the adversary's datasets in Table~\ref{tab:Datasets}.
Following our methodology, we begin by evaluating the extraction risk of naturally trained models, then of adversarially robust models. 

First, we train a \ac{DNN} on a given dataset. In our work, for both CIFAR-10 and SVHN we fine-tune a ResNet-34 \cite{he2016deep} pretrained model on ImageNet. For MNIST, we train a simple CNN architecture composed of two 2D-convolutional blocks, max pooling, dropouts and fully connected layers~\cite{pytorch_examples}.
In training, we use an SGD optimizer with an initial learning rate of 0.01 that is decreased by a factor of 10 every 60 epochs over up to 200 epochs \cite{atli2020extraction}. 
Then, we extract the victim model using the extraction attack KnockoffNets \cite{orekondy2019knockoff}. We experiment with multiple query budgets for up to 50000 samples.
We use as an adversary dataset, to query both CIFAR-10 and SVHN victims, a subset of images from ImageNet resized to $32\times32$ pixels using bilinear interpolation to match our input shapes for both victims networks. For the MNIST victim extraction, we leverage the FashionMNIST dataset (Table \ref{tab:Datasets}). 

Using the victim's prediction \emph{probabilities} as labels for the adversary's dataset, we train our surrogate models with a different architecture. We steal ResNet-34 victims with VGG-16 \cite{simonyan2014very} and the MNIST CNN with LeNet architecture \cite{Lecun1998lenet}. Surrogate models are trained using an SGD optimizer with an initial learning rate of 0.01 that is decreased by a factor of 10 every 60 epochs over up to 100 epochs.
Next, we evaluate the extraction success for each of these attacks. 

After that, we assess the extraction of adversarially robust models. Therefore, we take the training dataset and craft adversarial examples using one of the state-of-the art techniques in order to use them for retraining. In our work, we use two techniques Fast Gradient Sign Method (FGSM) \cite{goodfellow2014explaining} and Projected Gradient Descent (PGD) \cite{madry2017towards}  with 8 different perturbation levels $\varepsilon \in [0.01, 0.3]$.
Fig. \ref{fig:adversarial cifar10} illustrates some adversarial examples generated for the CIFAR-10 dataset using the PGD attack.

Using the generated adversarial examples from one of the techniques for a selected perturbation levels $\varepsilon \in [0.01, 0.15]$, where the noise added to the images does not fully deteriorate its visual perception, we augment the training set and retrain our models using the same training hyperparameters for the naturally trained victims. We obtain 8 \emph{adversarially robust} models for each $\varepsilon$ with PGD and FGSM techniques. Then, we perform the same extraction attack procedure on these models.
Finally, we compare the success of each attack on natural models versus their adversarially robust versions.

All experiments were performed on a Ubuntu 20.04 operating system with 8-core processor (3.7GHz $\times8$) and a GPU NVIDIA Quadro RTX 6000. We used Weights \& Biases (WandB)\cite{wandb} for experiment tracking and visualizations to develop insights for this paper~\footnote{Our project runs and visualizations are available at the WandB link: 
% \url{
https://wandb.ai/kacem/MSF
% }
% Link omitted for blind review.
}.

\subsection{Results}
% \subsection{Results}

\begin{table}[t]
\centering
\caption{Results of extraction attacks with selected budgets (B) against naturally trained DNN and adv. trained DNNs with different levels of $\varepsilon$ (Dataset: SVHN). The best accuracy and agreement of extracted models for each budget B are in  \textbf{bold}.} 
\begin{tabular}{|ll|c|lll|}
\hline
\multicolumn{1}{|c}{\multirow{2}{*}{\begin{tabular}[c]{@{}c@{}}Victim\\ Model\end{tabular}}} & \multicolumn{1}{c|}{\multirow{2}{*}{$\varepsilon$}} & \multirow{2}{*}{\begin{tabular}[c]{@{}c@{}}Test\\ Acc.\end{tabular}} & \multicolumn{3}{c|}{\begin{tabular}[c]{@{}c@{}}Extracted Models\\ Accuracy \& Agreement\end{tabular}} \\
\multicolumn{1}{|c}{} & \multicolumn{1}{c|}{} &  & \multicolumn{1}{c}{B=15k} & \multicolumn{1}{c}{B=25k} & \multicolumn{1}{c|}{B=50k} \\
\hline \hline
Natural & \multicolumn{1}{c|}{-} & 96.14 & 57.17 57.50 & 71.65 72.01 & 84.29 84.84 \\ \hline
\multirow{4}{*}{\begin{tabular}[c]{@{}l@{}}Adv.\\ trained\\ FGSM\end{tabular}} 
 & 0.03 & 96.33 & 59.36 59.73  & 72.08 72.62 & 85.69 86.29 \\
 & 0.05 & 96.54 & 61.07 61.42  & 76.34 76.79 & 86.46 86.92 \\
 & 0.10 & 96.41 & 62.91 63.35  & \textbf{78.12 78.74}  & 86.36 86.97 \\
 & 0.15 & 96.37 & 61.48 61.85  & 75.52 76.03  & 84.07 84.62 \\ \hline 
\multirow{4}{*}{\begin{tabular}[c]{@{}l@{}}Adv.\\ trained\\ PGD\end{tabular}} 
 & 0.03 & 96.53 & 59.36 59.73  & 72.08 72.62 & 84.57 85.17 \\
 & 0.05 & 96.40 & 63.09 63.37  & 75.21 75.59  & \textbf{86.76 87.37} \\
 & 0.10 & 96.47 & \textbf{65.39 65.83}  & 76.26 76.76  & 85.72 86.28 \\
 & 0.15 & 96.42 & 62.76 63.34  & 74.81 75.29  & 84.81 85.54 \\
 \hline
\end{tabular}
\label{tab:extraction-svhn}
\end{table}

\begin{figure}[tb]
\centering
\subfloat[Dataset: SVHN\label{fig:extraction-svhn-budget}]{%
  \includegraphics[width=0.499\textwidth]{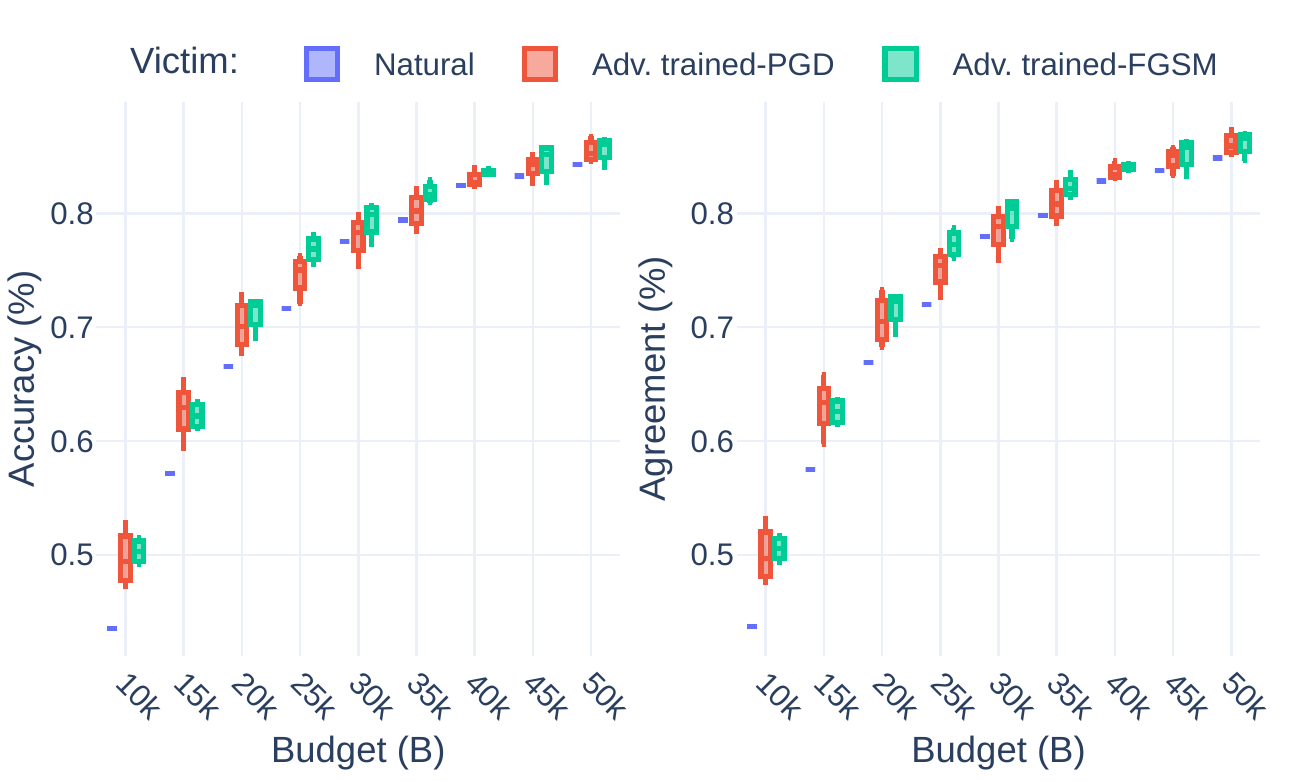}%
} \hfil
\subfloat[Dataset: CIFAR10\label{fig:extraction-cifar10-budget}]{%
  \includegraphics[width=0.499\textwidth]{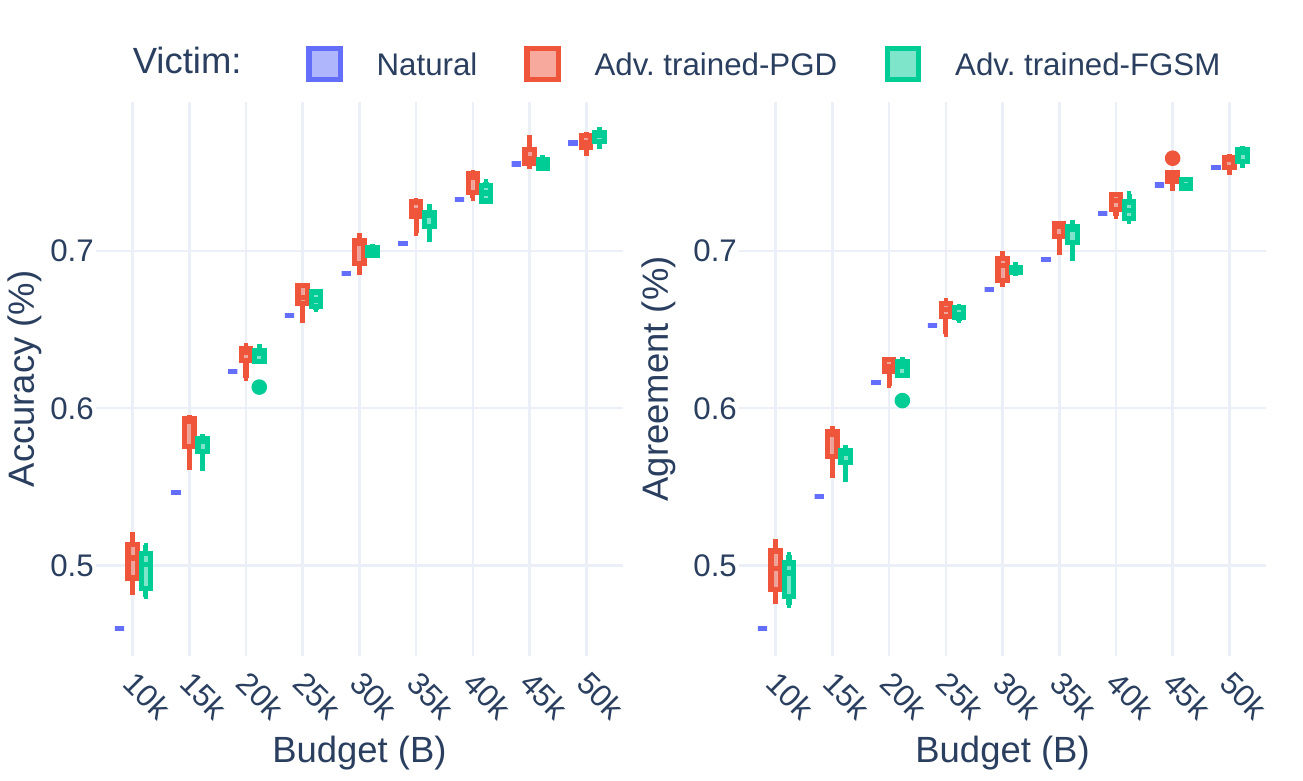}%
} \hfil
\subfloat[Dataset: MNIST\label{extraction-mnist}]{%
  \includegraphics[width=0.499\textwidth]{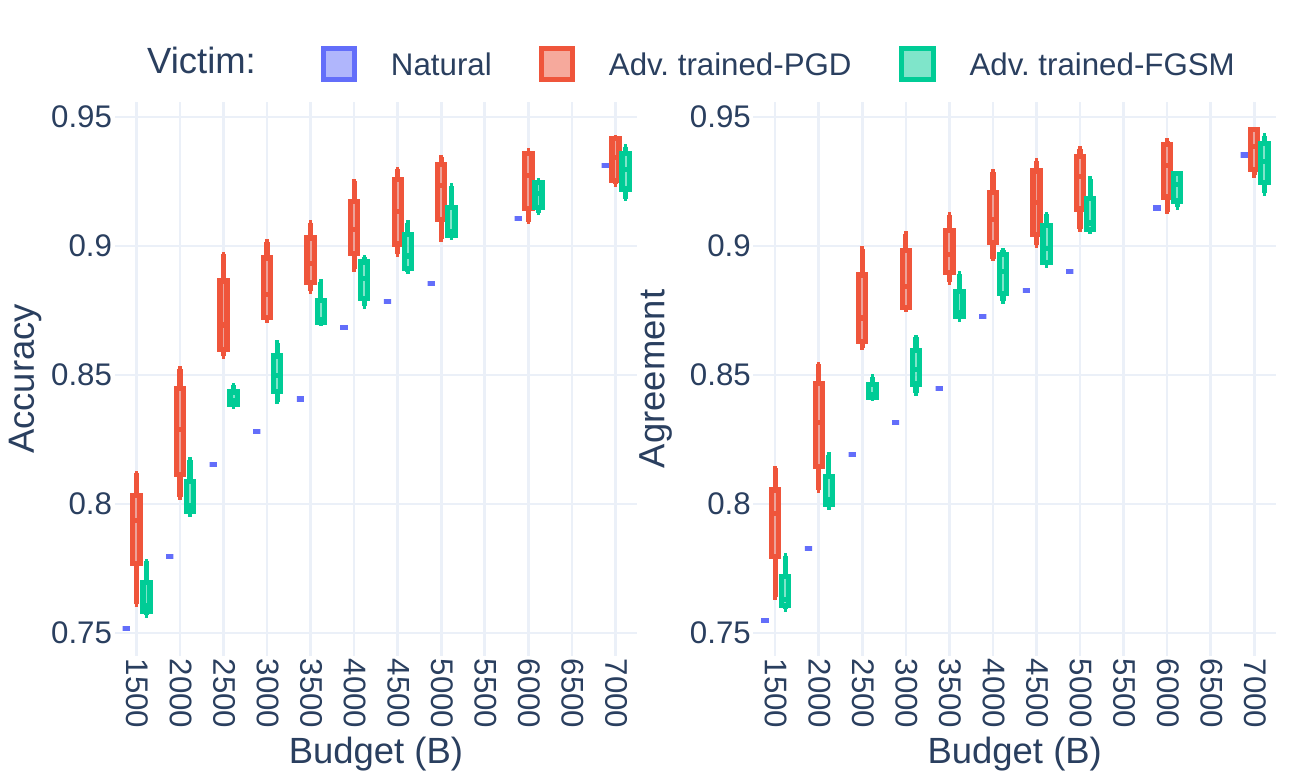}%
}
\caption{Impact of the adversarial training on the extraction attack success. For each victim model type, multiple extraction attacks are performed with different numbers of queries (B). The adv. trained models from FGSM and PGD results are grouped in box plots. For each technique, the boxes present results from victims with different levels of $\varepsilon \in [0.03, 0.15]$ that were used in the adversarial training.}
\label{fig:extraction-budget}
\end{figure}

\begin{figure*}[]
\centering
\subfloat[Dataset: SVHN\label{acc_adv-svhn}]{%
  \includegraphics[width=0.499\textwidth]{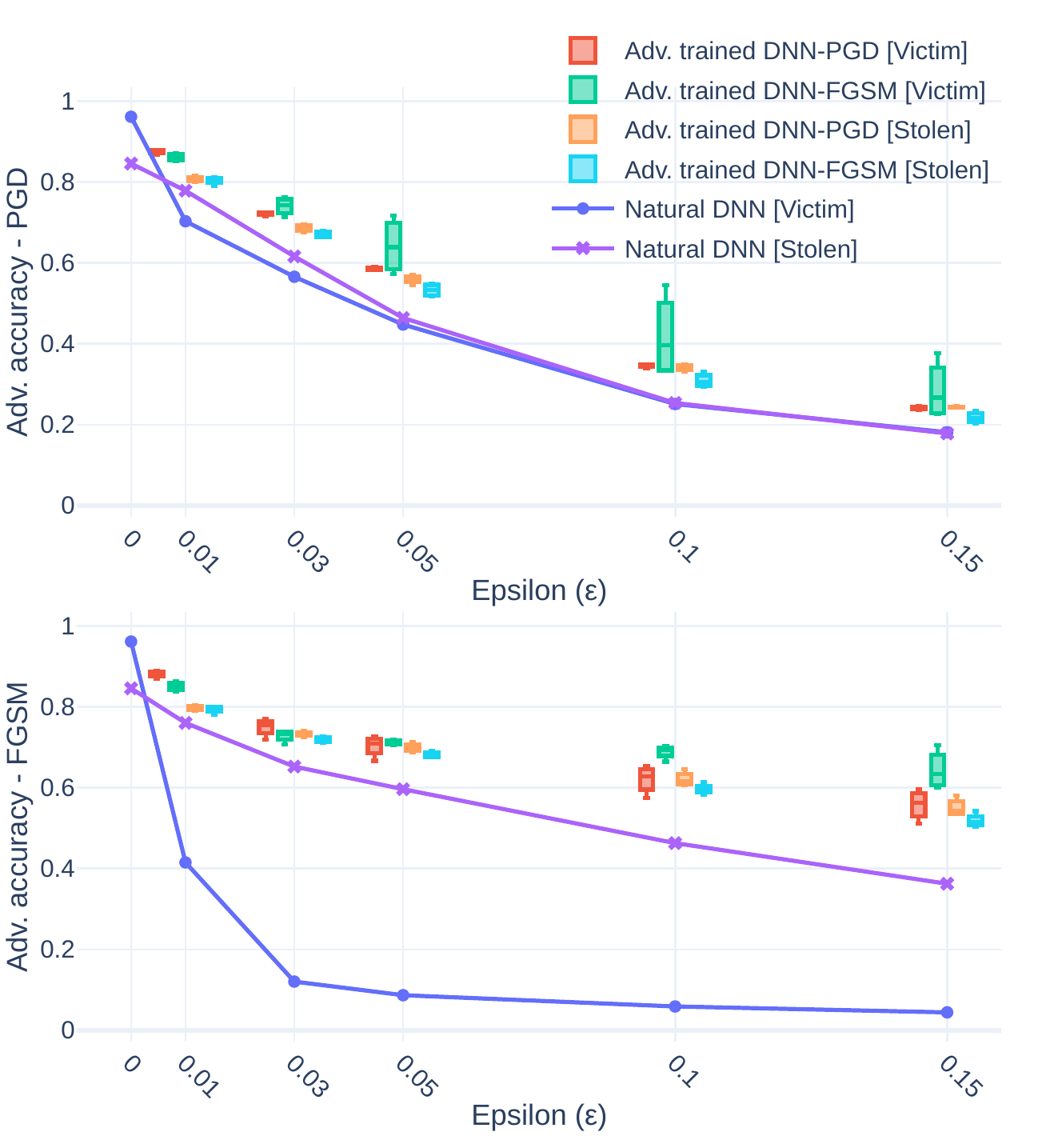}%
}\hfil
\subfloat[Dataset: CIFAR10\label{acc_adv-cifar10}]{%
  \includegraphics[width=0.499\textwidth]{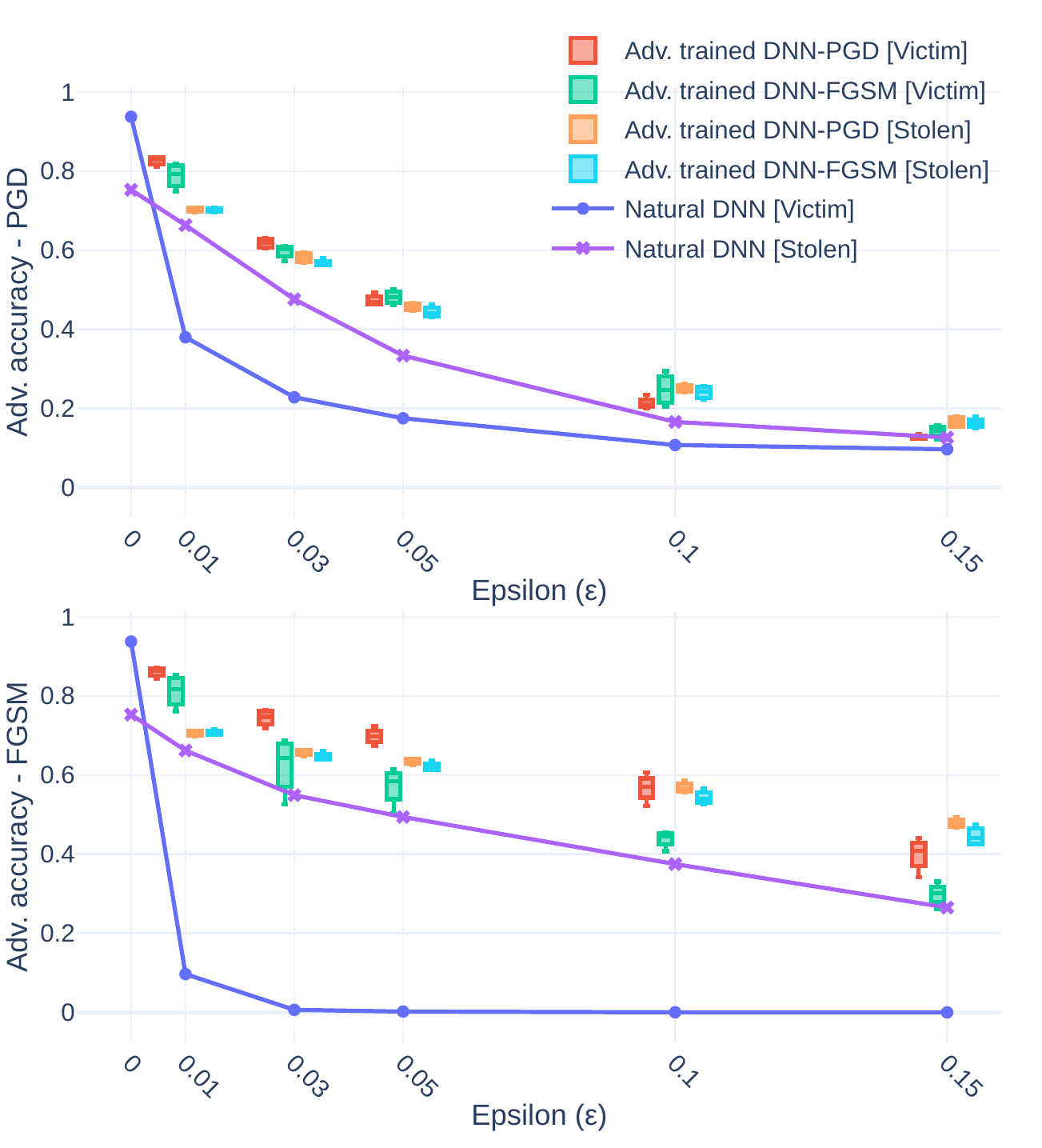}%
}
\caption{Adversarial robustness of victims and corresponding stolen models against multiple sets of adversarial examples with different $\varepsilon$ levels. 
The top row shows the accuracy against adversarial examples generated with PGD attack, while the second row concerns the FGSM attack. 
For each victim model type, we select the corresponding stolen model obtained with the highest budget in the performed extraction attack.
The adv. trained models from FGSM and PGD trained with different $\varepsilon$ levels are grouped in box plots as well as each of their corresponding stolen models.}
\label{fig:acc_adv}
\end{figure*}

\begin{figure}[]
\centering
% \subfloat[Dataset: SVHN\label{fig:extraction-svhn-eps}]{%
%   \includegraphics[width=0.22\textwidth]{figs/extraction-box-eps-SVHN.pdf}%
% }
% \hfil
% \subfloat[Dataset: CIFAR10\label{fig:extraction-cifar10-eps}]{%
%   \includegraphics[width=0.22\textwidth]{figs/extraction-box-eps-cifar10.pdf}%
% }
% \hfil
% \subfloat[Dataset: MNIST\label{extraction-gain-mnist}]{%
%   \includegraphics[width=0.22\textwidth]{figs/extraction-box-eps-MNIST.pdf}%
% }
\includegraphics[width=0.41\textwidth]{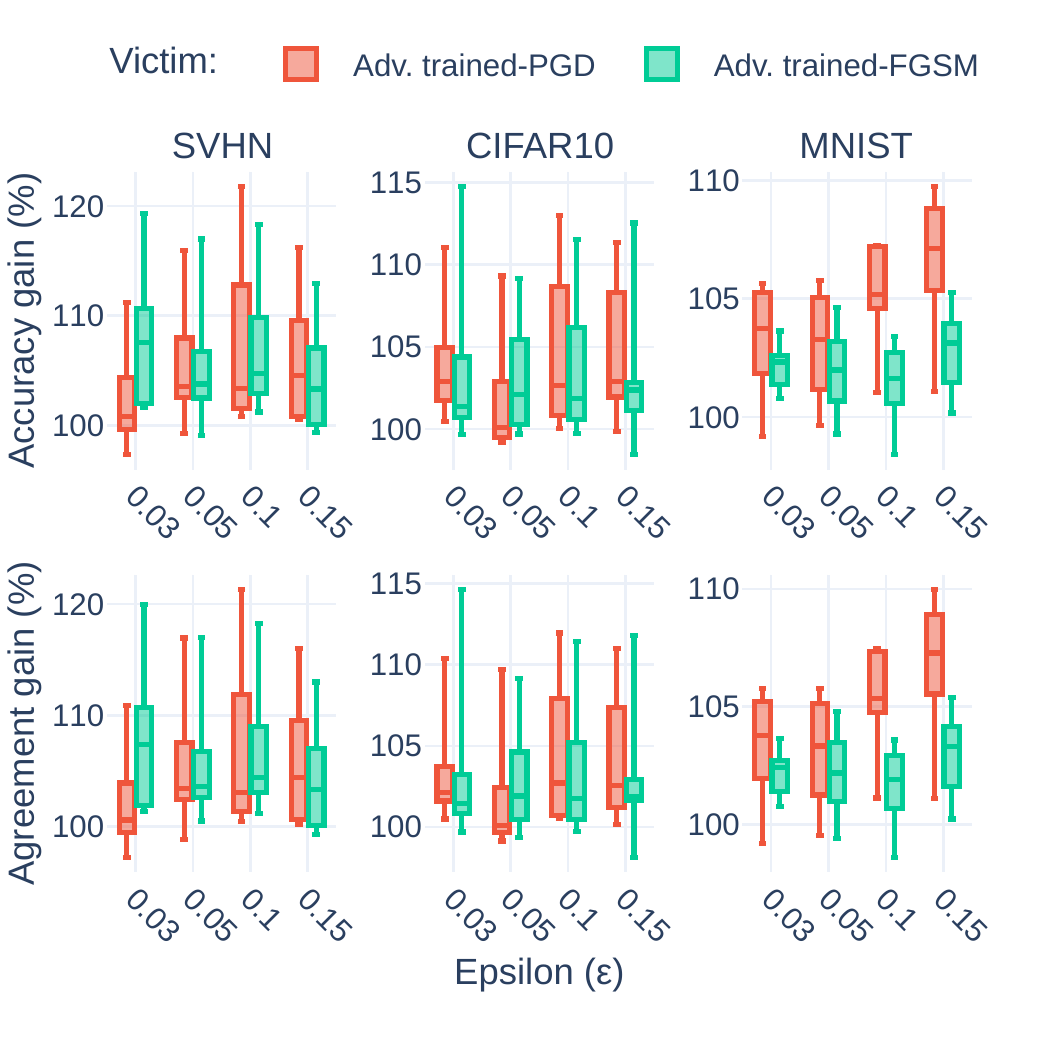}

\caption{Accuracy and agreement gains for the extraction of adversarially trained models compared to naturally trained models for different $\varepsilon$ levels. The adv. trained models from FGSM and PGD results are grouped in box plots. For each technique, the boxes present results from multiple extraction attacks which were performed with different numbers of queries (B).}
\label{fig:extraction-gain-eps}

\end{figure}

\subsubsection{Extraction of natural and adversarially trained models}
Table~\ref{tab:extraction-svhn} shows detailed results about the extraction attacks on naturally trained and adv. trained models on SVHN dataset. 
Each adv. trained victim is either an FGSM or PGD Adv. trained model with a specific $\varepsilon \in [0.03, 0.15]$. Each victim is extracted using a set of queries budget (B). 
We visualize  the \textit{accuracy} and the \textit{agreement} results of stolen models from different types of victims for SVHN, CIFAR-10 and MNIST datasets in Fig.~\ref{fig:extraction-budget}. 
We observe that adv. trained models have higher accuracy and higher agreement than naturally trained models. This gap is more noticeable with smaller budgets. This entails that adv. trained models are faster to extract compared to naturally trained models, i.e., adv. trained models require less number of queries in a stealing attack to have a higher extraction results. 

We quantify in Fig.~\ref{fig:extraction-gain-eps} the accuracy and agreement gains for the extraction of adversarially trained models with respect to naturally trained models depending on the noise level $\varepsilon$ that we use to generate adversarial examples for adversarial training. 
For both types of adv. trained models with different levels of~$\varepsilon$, in most cases the average gain in accuracy and agreement is higher than a naturally trained model. We find that surrogate models from the adv. trained models reach up to $\times1.2$ the extraction metrics of models stolen from a normally trained victim.

\subsubsection{Adversarial robustness of victims and surrogate models}
We evaluate the robustness of all models against several sets of adversarial examples  with different levels of $\varepsilon$. Fig.~\ref{fig:acc_adv} shows the adversarial accuracy of each model.
Note that naturally trained models are the most vulnerable to evasion attacks. 
We discover that an extracted model have a higher robustness against adversarial examples. 
The adversarial training of our victims is verified to be effective as a defense against evasion attacks as it shows an exceedingly better adversarial accuracy compared to naturally trained victims. 
Additionally, we discover that stolen models from adv. trained victims are as robust as the target models against adversarial examples and their adv. accuracy is higher than one of a stolen model from a natural victim. 
This demonstrates that the capability to be robust against adversarial examples  is transferable through extraction attacks.

\subsection{Discussion}
Our evaluation demonstrates that defending against adversarial examples through adversarial training may increase the vulnerability of models against extraction attacks. This proves that following a security-imposed scenario, the privacy of models can be jeopardized through stealing attacks.  

In addition, our work finds that stealing adversarially trained models rewards the thief with a surrogate model that is robust to adversarial examples. In fact, this finding may be explained by the generalization capacity of surrogate models. Since stolen models are trained by images from outside the distribution of the attacked victim's train set, the adversarial examples, which were crafted to evade the classification of the victim, fool less the extracted models. Hence, this makes the latter more robust to adversarial examples.
This concept have some intersections with a prior defense against adversarial examples through knowledge distillation of a model to another one using a temperature $T$ \cite{Papernot2016distillation}. 
In fact, both distillation and extraction attacks rely on training a model with data labeled by another model. However, the defensive distillation technique relies on the same training data rather than a different out of distribution images. Besides, this technique is performed by the model owner (white-box settings), not a malicious user that aims to steal the model (black-box settings).

\section{Defenses and countermeasures}
\label{sec:defenses}

Model extraction attacks have several drawbacks on \ac{ML} security and privacy, therefore, researchers propose different techniques to defend against them and mitigate their risk. 
Some techniques suggest watermarking the model to help claim it when stolen \cite{szyller2019dawn, jia2020entangled}, for example by changing the output probabilities for a small subset of queries (e.g., $\leq 0.5\%$) from API clients which allows model owners to reliably demonstrate ownership.
Other defenses can be broadly categorized into two main categories: \emph{proactive} and \emph{reactive} techniques. 
The former includes techniques to reduce efficiency of an extraction attack such as changing the prediction probabilities \cite{lee2018defending,kariyappa2020defending, Orekondy2019poisoning}. 
But, they may sometimes degrade the model accuracy and result in inference delays and an increase in computational costs.
Reactive defenses are based on the detection of extraction attacks when they are happening which enables the victim to take action immediately. 
These techniques continually observe the API query and response streams of users in order to either train a substitute model to assess the knowledge gained by the adversaries~\cite{Kesarwani2018},
or analyse the distribution of successive users queries to identify a deviation from a normal (Gaussian) distribution~\cite{Juuti2019} or a large number of out of distribution queries~\cite{kariyappa2020defending}.
However, these defenses do not detect all the attacks and they often consume more computational power.

\section{Conclusion}
\label{sec:conclusion}
% Conclusion

Despite the tremendous progress of neural networks and their trending and wide use in critical applications such as robotics and autonomous vehicles, they are still showing flaws and vulnerabilities in both security and privacy.
% Prior researchers evaluate different potential threats separately from each other, which raises questions about the robustness under these .
%
We have systematically explored model privacy threats through extraction attacks on a security-imposed scenario of adversarially trained models on three benchmark vision datasets.
We discovered that adversarially trained models might have a higher extraction rate compared to natural models for a lower number of queries. 
% They could achieve up to $\times1.2$ the same accuracy and agreement (i.e., fidelity) of natural models with a fraction less than $\times0.75$ of the same queries. 
This calls for a required attention from model owners when  improving the security of their models against adversarial examples through adversarial training.
We additionally found that extracted \ac{DNNs} from adv. trained models show an enhanced robustness against adversarial examples compared to extracted \ac{DNNs} from naturally trained models.

\bibliography{document}
\bibliographystyle{IEEEtran}			% Bibliography style. 

% \ifthenelse{\equal{\AnnexesPresentes}{O}}{
% 	\appendix%
% 	\newcommand{\Annexe}[1]{\annexe{#1}\setcounter{figure}{0}\setcounter{table}{0}\setcounter{footnote}{0}}%
% 	%\include{9-Annexes}
% 	}
% {}

\end{document}